\documentclass[11pt,a4paper]{article}
\usepackage[hyperref]{acl2020}
\usepackage{times}
\usepackage{latexsym}

\usepackage{graphicx}

\usepackage{microtype}

\usepackage{comment}
\usepackage{amsmath,amssymb} 
\usepackage{xcolor}

\usepackage{booktabs}
\usepackage{subcaption}
\usepackage{multirow}

\aclfinalcopy 

\title{UniVL: A Unified Video and Language Pre-Training Model for Multimodal Understanding and Generation} 

\author{
	Huaishao Luo\textsuperscript{1}\thanks{~This work was done during the first author's internship in MSR Asia}~, 
	Lei Ji\textsuperscript{2,3,4}, 
	Botian Shi\textsuperscript{5}, 
	Haoyang Huang\textsuperscript{2}, \\
	{\bf Nan Duan\textsuperscript{2}, 
	Tianrui Li\textsuperscript{1}, 
	Jason Li\textsuperscript{6},
	Taroon Bharti\textsuperscript{6}, 
	Ming Zhou\textsuperscript{2}} \\
  \textsuperscript{1}Southwest Jiaotong University, Chengdu, China\\
  \textsuperscript{2}Microsoft Research Asia, Beijing, China \\ 
  \textsuperscript{3}Institute of Computing Technology, Chinese Academy of Science, Beijing, China  \\
  \textsuperscript{4}University of Chinese Academy of Sciences, Beijing, China  \\
  \textsuperscript{5}Beijing Institute of Technology, Beijing, China  \\
  \textsuperscript{6}Microsoft STCA, Beijing, China  \\
	{\tt huaishaoluo@gmail.com, leiji@microsoft.com}
}

\begin{document}
	\maketitle
	\begin{abstract}
		With the recent success of the pre-training technique for NLP and image-linguistic tasks, some video-linguistic pre-training works are gradually developed to improve video-text related downstream tasks. However, most of the existing multimodal models are pre-trained for understanding tasks, leading to a pretrain-finetune discrepancy for generation tasks. This paper proposes UniVL: a \textbf{Uni}fied \textbf{V}ideo and \textbf{L}anguage pre-training model for both multimodal understanding and generation. It comprises four components, including two single-modal encoders, a cross encoder, and a decoder with the Transformer backbone. Five objectives, including video-text joint, conditioned masked language model (CMLM), conditioned masked frame model (CMFM), video-text alignment, and language reconstruction, are designed to train each of the components. We further develop two pre-training strategies, stage by stage pre-training (StagedP) and enhanced video representation (EnhancedV), to make the training process of the UniVL more effective. The pre-train is carried out on a sizeable instructional video dataset HowTo100M. Experimental results demonstrate that the UniVL can learn strong video-text representation and achieves state-of-the-art results on five downstream tasks.
	\end{abstract}
	
	\section{Introduction}
	With the recent advances of self-supervised learning, pre-training techniques play a vital role in learning visual and language representation. The paradigm is to pre-train the model on a large scale \emph{unlabeled} data and fine-tune the downstream tasks using task-specific \emph{labeled} data. Inspired by the BERT \cite{devlin2019bert} model's success for NLP tasks, numerous multimodal image-language pre-training models \cite{lu2019vilbert,li2019unicoder,li2019visualbert} have been proposed. Their results have demonstrated the effectiveness of pre-training on various visual and language tasks such as visual question answering. Different from previous text pre-training or image-language pre-training, we focus on video-linguistic pre-training in this paper.
	\begin{figure}[tp] 
		\centering
		\includegraphics[width=0.48\textwidth]{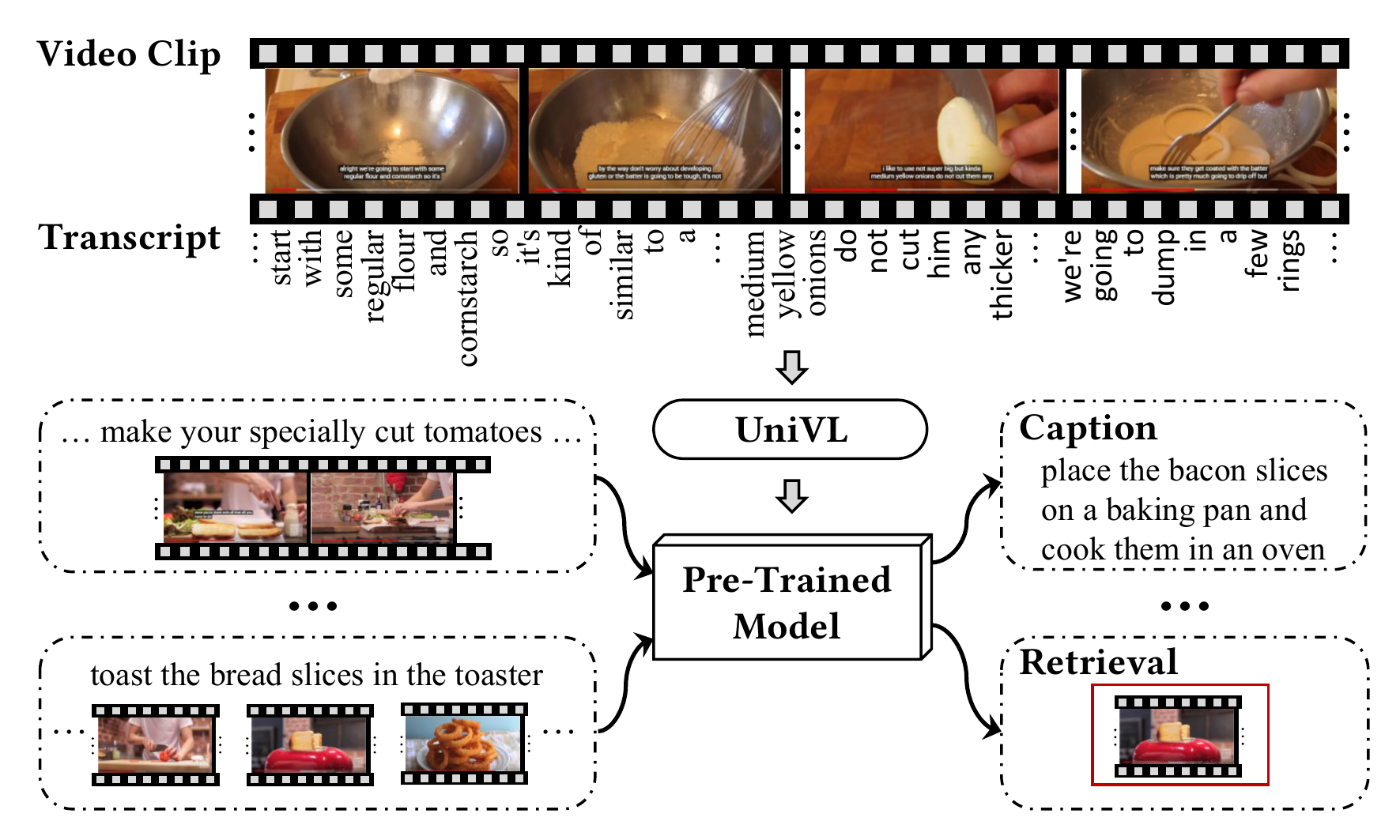} 
		\caption{A showcase of video and language pre-train based model for multimodal understanding (e.g., retrieval) and generation (e.g., captioning).}
		\label{fig:showcase}
	\end{figure}
	
	Videos contain rich visual, acoustic, and language information for people to acquire knowledge or learn how to perform a task. This motivates researchers to investigate whether AI agents can learn task completion from videos like humans with both low-level visual and high-level semantic language signals. Therefore, multimodal video-language tasks are of great importance to investigate for both research and applications. In this work, we first propose to pre-train a unified video-language model using video and acoustic speech recognition (ASR) transcript in instructional videos to learn a joint representation of both video and language. Then, we fine-tune this model on five typical multimodal tasks, including understanding and generation targets. Figure \ref{fig:showcase} presents a showcase of our pre-training and fine-tuning flow. Take multimodal video captioning as an example. The model inputs video and ASR transcript and predicts a captioning sentence.
	
	VideoBERT \cite{sun2019videobert} and CBT \cite{sun2019contrastive} are the first pioneers to investigate video-language pre-training with regard to video representation on instructional videos. They have demonstrated the effectiveness of the BERT based model for capturing video temporal and language sequential features. Besides the above two works, there is some concurrent progress to our model. ActBERT \cite{Zhu_2020_CVPR} leverages global action information to catalyze mutual interactions between linguistic texts and local regional objects. Moreover, a  transformer block is introduced to encode global actions, local regional objects, and linguistic descriptions. HERO \cite{Li2020HERO} hierarchically encodes multimodal inputs. Furthermore, two new pre-training tasks, video-subtitle matching and frame order modeling, are designed to improve the representation learning. VideoAsMT \cite{Korbar2020} takes a generative modeling approach that poses the objective as a translation problem between modalities.

	However, most of previous models only pre-train the model on understanding tasks. In this paper, we pre-train on both understanding and generation tasks through an encoder-decoder paradigm. Although the concurrent work VideoAsMT has a similar encoder-decoder as ours, it is not flexible for downstream tasks with only one single unified framework. In this paper, we develop a flexible approach to learn video and language joint representation and adapt downstream multimodal tasks.
	
	We propose UniVL: a \textbf{Uni}fied \textbf{V}ideo and \textbf{L}anguage pre-training model for multimodal understanding and generation. Our UniVL model adopts Transformer \cite{vaswani2017attention} as the backbone and has four components, including two single-modal encoders, a cross encoder, and a decoder. In detail, we first encode the text and visual separately by two single-modal encoders. A video-text joint objective performs on these two encoders, which aims to learn better representation for each modality before fusing them. Such a two-stream design is natural to retrieval tasks due to its scalability to very large datasets. The proposed representation can be indexed and has linear complexity in the number of videos. Then we adopt the Transformer based encoder-decoder model to perform the understanding and generation pre-training by four tasks: conditioned masked language model (CMLM for language corruption), conditioned masked frame model (CMFM for video corruption), video-text alignment, and language reconstruction.

	Furthermore, we design two pre-training strategies, including stage by stage pre-training strategy (StagedP) and Enhanced video representation (EnhancedV), to promote the UniVL pre-training. The StagedP has two parts in our setting. We only pre-train the text encoder and video encoder by the video-text joint objective for the first stage. Then all modules will be pre-trained under the whole objectives in the second stage. Besides, we adopt an entire masking strategy EnhancedV on text to enhance video representation.
	\begin{figure*}[tp]
		\centering
		\includegraphics[width=1.0\linewidth]{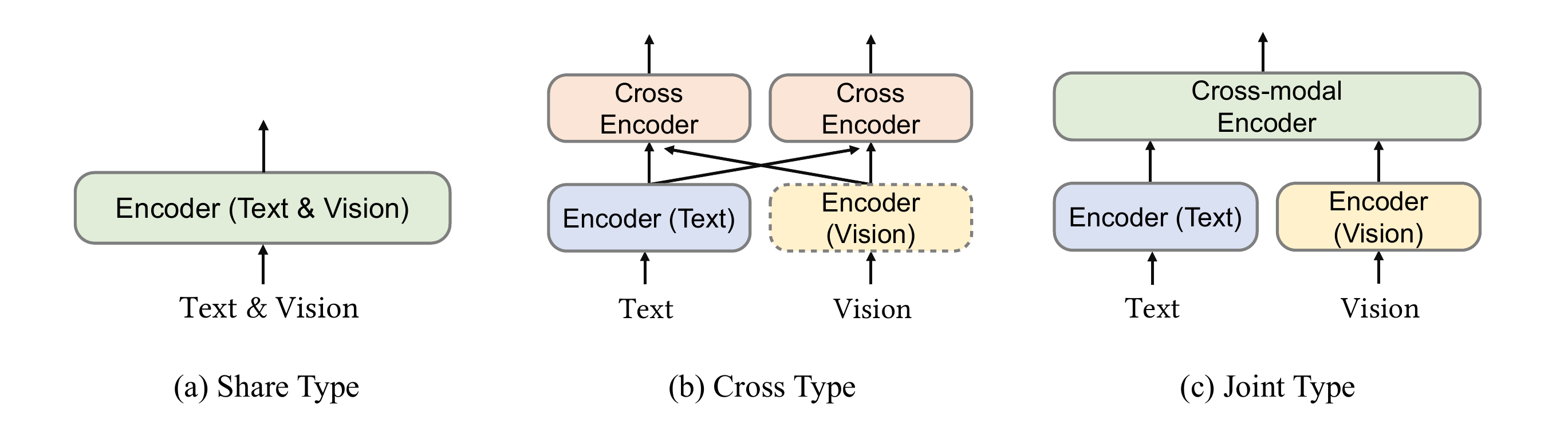}
		\caption{Various paradigms for multimodal pre-training.}
		\label{fig:paradigms}
	\end{figure*}
	
	Our contributions are summarized as follows:

	1) We propose a multimodal video-language pre-training model trained on a large-scale instructional video dataset. It is a flexible model for both video-language understanding and generation tasks. 

	2) The pre-training consists of five objectives, including video-text joint, conditioned masked language model, conditioned masked frame model, video-text alignment, and language reconstruction. Two pre-training strategies are proposed to make these objectives work harmoniously.

	3) We fine-tune our pre-trained model on five typical multimodal video-language tasks: text-based video retrieval, multimodal video captioning, action segmentation, action step localization, and multimodal sentiment analysis. Extensive experiments demonstrate our model's effectiveness on downstream tasks and achieve state-of-the-art results.
	
	\section{Related Works}
	\subsection{Single Modal Pre-Training}
	Self-supervised representation learning has been shown to be effective for sequential data, including language and video. Language pre-training models, including BERT \cite{devlin2019bert}, GPT \cite{radford2018improving}, RoBERTa \cite{liu2019roberta}, XLNet \cite{yang2019xlnet}, MASS \cite{song2019mass}, UniLM \cite{dong2019unified}, and BART \cite{lewis2019bart}, have achieved great success on NLP tasks. BERT \cite{devlin2019bert} is a denoising auto-encoder network using Transformer with MLM (masked language model) and NSP (next sentence prediction) as pre-training tasks. It has a strong performance for understanding tasks. MASS \cite{song2019mass} focuses on pre-training for generation tasks. UniLM \cite{dong2019unified} and BART \cite{lewis2019bart} continuously study a unified pre-training model for both understanding and generation tasks.
	
	Video representation learning mostly focuses on the video sequence reconstruction or future frames prediction as pre-training (pretext) tasks. Early works like \cite{mathieu2015deep,srivastava2015unsupervised,han2019video} aim to synthetic video frames through the image patches. Similarly, \citet{wang2015unsupervised} adopt a siamese-triplet network to rank continuous patches more similar than patches of different videos. Other works predict the feature vectors in latent space using auto-regressive models with the noise-contrastive estimation (NCE) \cite{lotter2016deep,oord2018representation}. \citet{sun2019contrastive} adopt NCE to predict corrupted (masked) latent space using the auto-encoder model.
	
	\subsection{Multimodal Pre-Training}
	Recently, numerous visual-linguistic pre-training models are proposed for multimodal tasks. For image and text pre-training, ViLBERT \cite{lu2019vilbert}, LXMERT \cite{tan2019lxmert} adopt two separate Transformers for image and text encoding independently. Other models like Visualbert \cite{li2019visualbert}, Unicoder-VL \cite{li2019unicoder}, VL-BERT \cite{Su2020VLBERT}, UNITER \cite{zhou2019unified} use one shared BERT model. These models employ MLM and image-text matching as pre-training tasks which are effective for downstream multimodal tasks. VLP \cite{zhou2019unified} proposes a unified image-language model for understanding and generation tasks. Different from these works, we focus on video and text pre-training for universal representation.
	
	For video and text pre-training, VideoBERT \cite{sun2019videobert} and CBT \cite{sun2019contrastive} are the first works to explore the capability of pre-training models. Although VideoBERT and CBT pre-train the model on multimodal data, the downstream tasks mainly take video representation for further prediction. ActBERT \cite{Zhu_2020_CVPR} leverages global action information to catalyze mutual interactions between linguistic texts and local regional objects, and introduces a  transformer block to encode global actions, local regional objects, and linguistic descriptions. HERO \cite{Li2020HERO} encodes multimodal inputs in a hierarchical fashion. Besides, two new pre-training tasks, video-subtitle matching and frame order modeling, are designed to improve the representation learning. However, ActBERT and HERO are only pre-train the models on understanding tasks. VideoAsMT \cite{Korbar2020} takes a generative modeling approach that poses the objective as a translation problem between modalities. The difference between our work with VideoAsMT is that our model contains two more separate encoders instead of one unified encoder-decoder, while VideoAsMT is inflexible for downstream tasks due to one single unified framework. 
	
	We summarize three pre-training paradigms to cover the previous vision-text pre-training model considering different encoder architecture in literature, as presented in Figure \ref{fig:paradigms}. Unicoder-VL \cite{li2019unicoder}, VL-BERT \cite{Su2020VLBERT}, UNITER \cite{zhou2019unified}, VLP \cite{zhou2019unified}, VideoBERT \cite{sun2019videobert}, ActBERT \cite{Zhu_2020_CVPR}, and VideoAsMT \cite{Korbar2020} belong to share-type in Figure \ref{fig:paradigms}(a), where the text and vision sequences are combined as the input of one shared Transformer encoder. ViLBERT \cite{lu2019vilbert} and LXMERT \cite{tan2019lxmert} are cross-type shown in Figure \ref{fig:paradigms}(b). CBT \cite{sun2019contrastive} and HERO \cite{Li2020HERO} are joint-type shown in Figure \ref{fig:paradigms}(c). The cross-type and joint-type architectures have two-stream input, and the difference is the interaction across both modalities. Compared with the single-stream input in the share-type, the two-stream input can accommodate each modality's different processing needs and interact at varying representation depths \cite{lu2019vilbert}. Besides, the joint-type structure has one cross-modal encoder for full interaction between the two streams comparing with the cross-type. We adopt the joint-type as our encoder in this paper.
	
	\section{Method}
	The problem is defined as: given the input video and the corresponding ASR transcript pairs, pre-train a model to learn joint video and text representation with the self-supervision approach, and fine-tune downstream tasks. In this section, we describe the architecture and pre-training tasks in detail.
	
	\subsection{Model Architecture}
\begin{figure*}[tp]
		\centering
		\includegraphics[width=0.99\textwidth]{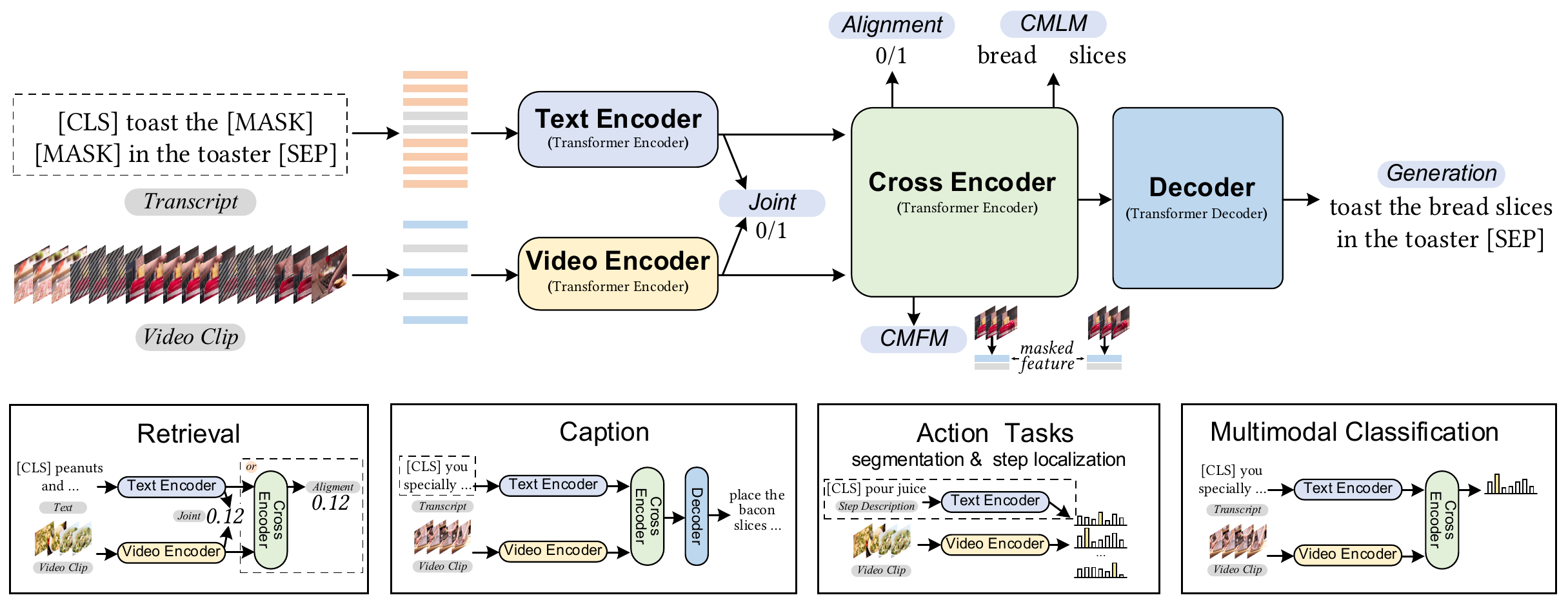} 
		\caption{The main structure of our UniVL, which comprises four components, including two single-modal encoders, a cross encoder, and a decoder. The model is flexible for many text and video downstream tasks. Four possible tasks are listed.}
		\label{fig:main_structure}
	\end{figure*}

	Figure \ref{fig:main_structure} presents the UniVL as an encoder-decoder architecture. First, the model extracts representations of the input text tokens and the video frame sequences using various feature extractors. A text encoder then adopts the BERT model to embed the text, and a video encoder utilizes the Transformer encoder to embed the video frames. Next, we employ a Transformer based cross encoder for interacting between the text and the video. Finally, a Transformer decoder is used to reconstruct the input text.
	
	\subsubsection{Pre-processing.} We first pre-process video and language before feeding to the UniVL. For the input text, we tokenize all words by WordPieces \cite{wu2016google} following the pre-processing method in BERT to obtain the token sequence $\mathbf{t}=\big\{t_i | i \in [1, n] \big\}$, where $t_i$ is the $i$-th token and $n$ is the length of the token sequence. For each video clip, we sample a frame sequence $\mathbf{v}=\big\{v_j | j \in [1, m] \big\}$ and adopt them to extract features, where $v_j$ is the $j$-th group of video frames and $m$ is the group length of the frame sequence.
	
	\subsubsection{Single Modal Encoders.} We encode the text and video separately. Such a two-stream design has two advantages: module reusing and retrieval orienting. The module reusing means the text module can benefit from the existing text-based pretrained model, e.g., BERT. The retrieval orienting means the two-stream design is natural to retrieval tasks due to its scalability to extensive datasets. The extracted representation can be indexed, and the calculation of similarity has linear complexity in the number of videos. In this paper, we adopt the BERT-Base uncased model to generate the  text representation $\mathbf{T} \in \mathbb{R}^{n \times d}$ after feeding the token sequence $\mathbf{t}$,
	\begin{align}
		\mathbf{T} = \text{BERT}(\mathbf{t}),
	\end{align}
	where $d$ is the hidden size of text representation. 
	
	For the video frame sequence $\mathbf{v}$, we adopt the off-the-shelf image feature extractors, e.g., S3D \cite{Xie2018Rethinking}, to generate video feature $\mathbf{F}_v \in \mathbb{R}^{m \times d^f_v}$, where $d^f_v$ is the hidden size. A Transformer encoder is utilized to embed the contextual information of video as follows,
	\begin{align}
		\mathbf{V} = \text{Transformer}(\mathbf{F}_v).
	\end{align}
	The dimension of $\mathbf{V}$ is $\mathbb{R}^{m \times d}$.
	
	\subsubsection{Cross Encoder.} The text encoder and video encoder mainly focus on individual modality. To make the text and video fully interact, we design across encoder, which takes both the text and video modality features as input. Specifically, we first combine the text encoding $\mathbf{T}$ and the video encoding $\mathbf{V}$ to get the encoding $\mathbf{M} \in \mathbb{R}^{(n+m) \times d}$. Then, a Transformer encoder takes the encoding $\mathbf{M}$ as input to generate the attended encoding $\mathbf{M} \in \mathbb{R}^{(n+m) \times d}$, 
	\begin{align}
		\mathbf{M} = \text{Transformer}(\left[\mathbf{T} ; \mathbf{V}\right]),
	\end{align}
	where $[;]$ denotes the combination operation. It is noted that the combination is operated along with the dimension of sequence, not the dimension of hidden size. One reason is that the text length $n$ and video clip length $m$ are always different. Another reason is that the semantic between text and video are not absolutely aligned. People are likely to describe an event after or before performing it in the video \cite{miech19endtoend}.
	
	\subsubsection{Decoder.} We empower our pre-trained model to have the capability of learning from and then benefiting for generation tasks by attaching a decoder, which is usually a unidirectional recurrent/attention model to generate tokens one by one. Such a decoder module is proved useful in text-based pre-training tasks, e.g., T5 \cite{2019t5} and BART \cite{lewis-etal-2020-bart}. It is noted that the decoder has a different target at different phases. The decoder learns to reconstruct the input text (e.g., transcripts) during pre-training because of no available text label. When fine-tuning, the decoder is used to generate results, e.g., video caption, where inputs transcripts and video and outputs caption. The input is the attended encoding $\mathbf{M}$ of text and video. We unexceptionally exploit Transformer to get the decoded feature $\mathbf{D} \in \mathbb{R}^{l \times d}$ from $\mathbf{M}$,
	\begin{align}
		\mathbf{D} = \text{Transformer}(\mathbf{M}),
	\end{align}
	where $l$ is the decoder length.
	
	\subsection{Pre-training Objectives}
	We have five pre-training objectives: 1) video-text joint, 2) conditioned masked language model (for text corruption), 3) conditioned masked frame model (for video corruption), 4) video-text alignment, and 5) language reconstruction.
	
	\subsubsection{Video-Text Joint.} As our text encoder, the BERT-Base uncased model is a robust extractor of text representation. So, we utilize a video-text joint objective to enhance the capability of the video encoder. It seems a retrieval orienting operation, which is to align the space of representation between text and video. Considering the misalignment between the text and video clip in narrated videos, we adopt MIL-NCE \cite{miech19endtoend} on $\mathbf{T}$ and $\mathbf{V}$ as our joint objective,
	\begin{align}
		&\mathcal{L}_{Joint}(\theta) = -E_{(\mathbf{t}, \mathbf{v}) \sim \mathbf{B}} \log \text{MIL-NCE}\left( \mathbf{t}, \mathbf{v} \right), \label{eq:joint_loss} \\
		&\text{MIL-NCE}\left( \mathbf{t}, \mathbf{v} \right) = \frac{ \sum_{(\mathbf{\hat{v}},\mathbf{\hat{t}}) \in \mathcal{P}_{\mathbf{v},\mathbf{t}}} {\exp(\mathbf{\hat{v}}\mathbf{\hat{t}}^{\top})} }{\mathcal{Z}}, \label{eq:dot_product} \\
		&\mathcal{Z} = \sum\limits_{(\mathbf{\hat{v}},\mathbf{\hat{t}}) \in \mathcal{P}_{\mathbf{v},\mathbf{t}}}{\exp(\mathbf{\hat{v}}\mathbf{\hat{t}}^{\top})} + \sum\limits_{(\mathbf{\tilde{v}},\mathbf{\tilde{t}}) \in \mathcal{N}_{\mathbf{v},\mathbf{t}}}{\exp(\mathbf{\tilde{v}}\mathbf{\tilde{t}}^{\top})},
	\end{align}
	where $\mathcal{P}_{\mathbf{v},\mathbf{t}}$ is a set of positive video-transcript pairs. E.g., $\{(\mathbf{v},\mathbf{t}), (\mathbf{v},\mathbf{t}_{-1}), (\mathbf{v},\mathbf{t}_{+1})\}$, where $\mathbf{t}_{-1}$ and $\mathbf{t}_{+1}$ are two closest transcripts in time to $\mathbf{t}$. The negative pairs $\mathcal{N}_{\mathbf{v},\mathbf{t}}$ take negative transcripts (or video clips) from other instances within the batch $\mathbf{B}$ after fixing $\mathbf{v}$ (or $\mathbf{t}$). $\mathbf{\hat{v}}, \mathbf{\tilde{v}}$ and $\mathbf{\hat{t}}, \mathbf{\tilde{t}}$ are generated through mean-pooling on $\mathbf{V}$ and $\mathbf{T}$, respectively.  $\theta$ is the trainable parameters.
	
	\subsubsection{CMLM: Conditioned Masked Language Model.} Following BERT, we also randomly mask 15\% tokens with the special token $\text{[MASK]}$ in the sentence and re-produce the masked tokens under the condition of video input and known tokens.  This loss function is defined on the feature matrix of the text part in $\mathbf{M}$ as: 
	\begin{align}
		\mathcal{L}_{CMLM}(\theta) = -E_{t_m \sim \mathbf{t}} \log P_{\theta}\left( t_m \mid t_{\neg m}, \mathbf{v} \right),
	\end{align}
	where $t_{\neg m}$ means the contextual tokens surrounding the masked token $t_m$, $\theta$ is the trainable parameters.
	
	\subsubsection{CMFM: Conditioned Masked Frame Model.} Similarly, we also propose a masked frame model to predict the correct frames given contextual frames and the input text for semantic constraints. However, it is hard to reconstruct the original RGB frame. We adopt the contrastive learning method to maximize the MI (Mutual information) between the masked output features and the original features. This loss function is NCE \cite{sun2019contrastive}. We randomly mask 15\% vectors (also 15\% frames) with zeros. The objective is to identify the correct frame compared to negative distractors. The loss is defined as:
	\begin{align}
		&\mathcal{L}_{CMFM}(\theta) = -E_{v_m \sim \mathbf{v}} \log \text{NCE}\left( v_m \mid v_{\neg m}, \mathbf{t} \right), \\
		&\text{NCE}\left( v_m \mid v_{\neg m}, \mathbf{t} \right) = \frac{\exp(\mathbf{f}_{v_m}\mathbf{m}_{v_m}^{\top})}{\mathcal{Z}}, \\
		&\mathcal{Z} = \exp(\mathbf{f}_{v_m}\mathbf{m}_{v_m}^{\top}) + \sum\nolimits_{v_j \in \mathcal{N}(v_m)}\exp(\mathbf{f}_{v_m}\mathbf{m}_{v_j}^{\top}),
	\end{align}
	where $v_{\neg m}$ means the surrounding frames except $v_m$, $\mathbf{f}_{v_m} \in \mathbb{R}^{1 \times d}$ is a linear output of $\mathbf{f}^{v}_{v_m} \in \mathbf{F}_v$, $\mathbf{F}_v$ is the real-valued vectors of video features, $\mathbf{m}_{v_m} \in \mathbf{M}^{(v)}$, and $\mathbf{M}^{(v)}$ is the feature matrix of the video part in $\mathbf{M}$. We take other frames in the same batch as negative cases defined as $\mathcal{N}(v_m)$.
	
	\subsubsection{Video-Text Alignment.} 
	We use the fused representation that corresponds to the special token $\text{[CLS]}$ to predict scores for the video-text alignment, which is similar to the BERT sentence pair classification task. We adopt the NCE loss to learn to discriminate against the positive from negative video-text pairs. To enhance this capability, we not only randomly sample negative cases but also re-sample video clips from the same video \cite{han2019video}. The reason is that the frames inside the same video are more similar than frames of different videos. This loss function is defined as follows,
	\begin{align}
		&\mathcal{L}_{Align}(\theta) = -E_{(\mathbf{t}, \mathbf{v}) \sim \mathbf{B}} \log \frac{\exp\big(s(\mathbf{t},\mathbf{v})\big)}{\mathcal{Z}}, \label{eq:loss_align} \\
		&\mathcal{Z} = \exp\big(s(\mathbf{t},\mathbf{v})\big) +\!\! \sum\nolimits_{\mathbf{u} \in \mathcal{N}(\mathbf{v})}\exp\big(s(\mathbf{t},\mathbf{u})\big),
	\end{align}
	where $s(\cdot)$ means two linear layers with a $Tanh$ activation function between them, which is performed on the first hidden state of $\mathbf{M}$. We take other video clips in the same batch $\mathbf{B}$ as negative cases $\mathcal{N}(\mathbf{v})$.
	
	\subsubsection{Language Reconstruction.} 
	To reconstruct the input sentence to endow the pretrained model with the generation capability, we employed an auto-regressive decoder with reconstruction objective, and the loss function is,
	\begin{align}
		\mathcal{L}_{Decoder}(\theta) = -E_{\hat{t}_i \sim \mathbf{\hat{t}}} \log P_{\theta}\left( \hat{t}_i \mid \hat{t}_{< i}, \mathbf{t}, \mathbf{v} \right).
	\end{align}
	It is noted that $\mathbf{t}$ is the masked version of ground-truth text $\mathbf{\hat{t}}$ when pre-training. As shown in BART \cite{lewis2019bart}, pre-training decoder benefits generation tasks. 
	
	We jointly optimize our model by a weighted loss:
	\begin{align}
		\mathcal{L}_{UniVL} = &\mathcal{L}_{Joint} + \mathcal{L}_{CMLM} + \mathcal{L}_{CMFM} \notag\\
		&+ \mathcal{L}_{Align} + \mathcal{L}_{Decoder}.
	\end{align}
	
	\subsection{Pre-training Strategies}
	We develop two pre-training strategies to train the UniVL model effectively.
	
	\subsubsection{StagedP: Stage by Stage Pre-training.} The UniVL can benefit from the pre-trained BERT-Base uncased model in the text encoder module. The natural idea is to train a peer to peer video encoder as the BERT-Base. We adopt a two-stage training fashion. For the first stage, we only preserve the text BERT and video Transformer to learn the weights using the Video-Text Joint loss Eq. (\ref{eq:joint_loss}). Next, we decrease the learning rate and continue to further pre-train the UniVL by all five objectives. One advantage is to fasten the pre-training speed, and the other advantage is to make the pre-training progress more smoothing on weights.
	
	\subsubsection{EnhancedV: Enhanced Video Representation.}
	To further enhance the video representation, we adopt a masked modality strategy to make the video to generate transcripts without text input. Specifically, we mask the whole text tokens with a 15\% possibility. In other words, there are 15\% text-video pairs with entire text tokens masked in each mini-batch, and the model utilizes the video information to complete generation. Such a strategy is a more challenging task for the model to learn a better video representation.
	
	\section{Experiments}
	We first pre-train our model on the large scale dataset. We download videos with ASR transcripts from Howto100M dataset \cite{miech2019howto100m}\footnote{https://www.di.ens.fr/willow/research/howto100m/}. After filtering the unavailable ones, we get 1.2M videos for pre-training our model. On average, the duration of each video is 6.5 minutes with 110 clip-text pairs.
	
	Then, we fine-tune our pre-trained model on five diverse downstream tasks using five datasets, including text-based video retrieval, multimodal video captioning, action segmentation, action step localization, and multimodal sentiment analysis.
	
	\subsection{Datasets}
	\subsubsection{Youcook2} Youcook2 \cite{zhou2018towards} contains 2,000 cooking videos on 89 recipes with 14K video clips. The overall duration is 176 hours (5.26 minutes on average). Each video clip is annotated with one captioning sentence. We evaluate both text-based video retrieval and multimodal video captioning task on this dataset. 
	
	For the text-based video retrieval task, we follow the same experimental setting in \cite{miech2019howto100m}, and use the captions as the input text queries to find the corresponding video clips. For the video captioning task, we use the same setting as in \cite{shi2019dense}. We filter the data and make sure there is no overlap between pre-training and evaluation data. In all, we have 1,261 training videos and 439 test videos, that is, 9,776 training clip-text pairs and 3,369 test clip-text pairs. 
	
	\subsubsection{MSR-VTT} MSR-VTT \cite{xu2016msr} is the open-domain dataset for video retrieval tasks. It has open domain video clips, and each clip has 20 captioning sentences labeled by human. In all, there are 200K clip-text pairs from 10K videos in 20 categories including sports, music, etc. Following JSFusion \cite{yu2018joint}, we randomly sampled 1,000 clip-text pairs as test data to evaluate the performance of our model on text-based video retrieval task.
	
	\subsubsection{COIN} COIN \cite{Tang2019COIN} is to evaluate action segmentation task, which contains 180 different tasks and 11,827 videos. Each video is labeled with 3.91 step segments. In total, the dataset contains videos of 476 hours, with 46,354 annotated segments. 
	
	\subsubsection{CrossTask} CrossTask \cite{Zhukov2019Cross} is to evaluate the action step localization task. It contains 83 different tasks and 4.7k videos. For each task, an ordered list of steps with manual descriptions are provided. 
	
	\subsubsection{CMU-MOSI} Multimodal Opinion Sentiment and Emotion Intensity \cite{zadeh2016multimodal} is sentence-level sentiment analysis and emotion recognition in online videos. CMU-MOSI contains 2,199 opinion video clips, each annotated with real-valued sentiment intensity annotations in the range [-3, +3]. We evaluate the performance of our model on multimodal sentiment analysis.
	\begin{table*}[ht]
		\setlength{\tabcolsep}{4pt}
		\centering
		\begin{tabular}{lcccc}
			\toprule
			Methods  & R@1   & R@5 & R@10 & Median R  \\ \midrule
			Random   & 0.03	& 0.15	& 0.3 & 1675 \\  
			HGLMM \cite{klein2015associating} & 4.6 & 14.3 & 21.6 & 75 \\  
			HowTo100M \cite{miech2019howto100m}   & 8.2 & 24.5 & 35.3 & 24 \\ 
			MIL-NCE \cite{miech19endtoend} & 15.1 & 38.0 & 51.2 & 10 \\ 
			ActBERT \cite{Zhu_2020_CVPR}    & 9.6 & 26.7 & 38.0 & 19 \\ 
			VideoAsMT \cite{Korbar2020}   & 11.6 & - & 43.9 & - \\ 
			\midrule 
			UniVL (FT-Joint)    & 22.2 & 52.2 & 66.2 & 5 \\
			UniVL (FT-Align)    & \textbf{28.9} & \textbf{57.6} & \textbf{70.0} & \textbf{4} \\
			\bottomrule
		\end{tabular}
		\caption{Results of text-based video retrieval on Youcook2 dataset.}
		\label{tab:result_of_retrieval_youcook}
	\end{table*}
	
	\subsection{Experimental Details}
	For text encoding, we apply WordPiece embeddings \cite{wu2016google} with a 30,000 token vocabulary to input to BERT model. We exploit the BERT-base model \cite{devlin2019bert} with 12 layers of Transformer blocks. Each block has 12 attention heads and the hidden size is 768.
	
	For video encoding, we first extract the 3D feature from video clips using the S3D model pretrained by \citet{miech19endtoend}. The basic visual feature can significantly affect the results from our preliminary experiments. The fps of the 3D feature extractor is 16 and the dimension is 1,024. We then employ Transformer Encoder with 6 layers to capture the sequential information on the 3D feature. Each block has 12 attention heads and the hidden size is 768.
	
	The model consumes the clip-text pairs. The maximal input tokens of text is 32 and the maximal number of video features is 48. For short sentence and clip, we concatenate contextual tokens and frames. For cross encoder and decoder, we use a 2 layers Transformer Encoder as the encoder and a 3 layer Transformer Decoder as the decoder with 12 heads and 768 hidden size. For generation task during the inference stage, we use the beam search with the size of 5. As previously mentioned, the generated sequence is the ground-truth input transcripts in the pre-training phase. Its target is to sequentially learn full information from the masked transcripts and video features.
	\begin{table*}[tp] 
		\setlength{\tabcolsep}{4pt}
		\centering
		\begin{tabular}{lcccc}
			\toprule 
			Methods         & R@1   & R@5 & R@10 & Median R  \\ \midrule
			Random   & 0.1 & 0.5 & 1.0 & 500 \\  
			C+LSTM+SA \cite{Torabi2016Learning}             & 4.2 & 12.9 & 19.9 & 55 \\  
			VSE \cite{kiros2014unifying}            & 3.8 & 12.7 & 17.1 & 66 \\  
			SNUVL \cite{Yu2016VideoCaptioning}             & 3.5 & 15.9 & 23.8 & 44 \\  
			\citet{Kaufman2017Temporal}             & 4.7 & 16.6 & 24.1 & 41 \\ 
			CT-SAN \cite{Yu2017End}             & 4.4 & 16.6 & 22.3 & 35 \\ 
			JSFusion \cite{yu2018joint}             & 10.2 & 31.2 & 43.2 & 13 \\ 
			HowTo100M \cite{miech2019howto100m}   & 14.9 & 40.2 & 52.8 & 9 \\ 
			MIL-NCE \cite{miech19endtoend} & 9.9 & 24.0 & 32.4 & 29.5 \\ 
			ActBERT \cite{Zhu_2020_CVPR}    & 8.6 & 23.4 & 33.1 & 36 \\ 
			VideoAsMT \cite{Korbar2020}   & 14.7 & - & 52.8 & - \\ 
			\midrule
			UniVL (FT-Joint)  & 20.6 & 49.1 & 62.9 & 6 \\
			UniVL (FT-Align)  & \textbf{21.2} & \textbf{49.6} & \textbf{63.1} & \textbf{6} \\
			\bottomrule
		\end{tabular}
		\caption{Results of text-based video retrieval on MSR-VTT dataset.}
		\label{tab:result_of_retrieval_MSR-VTT}
	\end{table*}
	
	We pre-train our model on 8 NVIDIA Tesla V100 GPUs. There are two sets of hyper-parameters considering the stage by stage pre-training strategy. In the first stage, the batch size is set to 600 and the model is trained 50 epochs for 1.5 days. In the second stage, the batch size is set to 48 and the model is trained 50 epochs for 12 days. We use the Adam optimizer \cite{kingma2014adam} with an initial learning rate of 1e-3 in the first stage and 1e-4 in the second stage, and employ a linear decay learning rate schedule with a warm-up strategy.
	\begin{table*}[tbp] 
		\setlength{\tabcolsep}{6pt}
		\centering
		\scalebox{0.9}{
			\begin{tabular}{l|c|ccccc}
				\toprule
				Methods                   & Input & B-3            & B-4            & M                       & R-L              & CIDEr             \\ \midrule
				Bi-LSTM \cite{zhou2018towards}       & V & -              & 0.87           & 8.15                    &  -               &   -               \\ 
				EMT \cite{zhou2018end}                      & V & -              & 4.38           & 11.55                   &  27.44           &   0.38            \\ 
				VideoBERT \cite{sun2019videobert}                 & V  & 6.80           & 4.04           & 11.01                   &  27.50           &   0.49            \\  
				CBT  \cite{sun2019contrastive}                     & V  & -              & 5.12           & 12.97                   &  30.44           &   0.64            \\ 
				ActBERT \cite{Zhu_2020_CVPR}    & V  & 8.66 & 5.41 & 13.30 & 30.56 & 0.65 \\ 
				VideoAsMT \cite{Korbar2020}   & V  & - & 5.3 & 13.4 & - & - \\ 
				AT \cite{hessel2019case} & T  & -              & 8.55           & 16.93 &  35.54           &   1.06   \\ 
				\midrule
				DPC  \cite{shi2019dense} & V + T  & 7.60           & 2.76           & 18.08 &  -               &   -               \\ 
				AT+Video \cite{hessel2019case} & V + T  & -              & 9.01           & 17.77                   &  36.65           &   1.12   \\ 
				\midrule
				UniVL         & V  & 16.46 & 11.17 & 17.57 & 40.09 & 1.27    \\ 
				UniVL         & T  & 20.32  & 14.70 & 19.39 & 41.10 & 1.51 \\ 
				UniVL        & V + T  & \textbf{23.87} & \textbf{17.35} & \textbf{22.35} &  \textbf{46.52}  &   \textbf{1.81}   \\ 
				\bottomrule
			\end{tabular}
		}
		\caption{The multimodal video captioning results on Youcook2 dataset. `V' means video and `T' means Transcript.}
		\label{tab:caption_result}
	\end{table*}

	\subsection{Main Results}
	\subsubsection{Text-based Video Retrieval.}
	Text-based video retrieval is defined to retrieve a relevant video/clip given an input text query. As shown in Figure \ref{fig:main_structure} (retrieval block), the model encodes the input text query and candidate video clips through the text encoder and video encoder respectively. Then calculate the matching scores using two different approaches: one is UniVL (FT-Joint), which calculates the score through dot product as in Eq. (\ref{eq:dot_product}), and use $\mathcal{L}_{Joint}$ as the loss during the fine-tuning stage; the other is UniVL (FT-Align), which feeds the encodings to both single encoders and the cross encoder to get unified representation and predict the match score through $s(\cdot)$ in Eq. (\ref{eq:loss_align}) on the first token `[CLS]'. During the fine-tuning stage, the loss is $\mathcal{L}_{Align}$. We use the Adam optimizer with an initial learning rate of 3e-5 and a batch size of 32 video-caption pairs for Youcook2, an initial learning rate of 5e-5 and a batch size of 128 video-caption pairs for MSR-VTT as hyper-parameters to fine-tune for 5 epochs. 
	
	We fine-tune our pre-trained model for text-based video retrieval task on both Youcook2 and MSR-VTT datasets. The evaluation metrics are Recall@n (R@n) and Median R. Tables \ref{tab:result_of_retrieval_youcook} and \ref{tab:result_of_retrieval_MSR-VTT} list the retrieval results of all baselines and our model on Youcook2 and MSR-VTT separately. We can see that our model achieves the best performance over all baselines to a large extent. We present several baseline methods with or without pre-training. Our model outperforms the Howto100M and VideoAsMT models pre-trained on the same dataset on all metrics. Besides, the experimental results present the a large performance gain with pre-training.
	
	We also notice that UniVL (FT-Align) performs better than UniVL (FT-Joint), which demonstrates that fusion representation generated by the cross encoder is better. Nevertheless, the UniVL (FT-Joint) inference speed is 50 times for Youcook2 and 10 times for MSR-VTT faster than that of the UniVL (FT-Align). Therefore, it is a trade-off between performance and efficiency in practical applications. In the following ablation experiment, we exploit UniVL (FT-Joint) in the retrieval task.
	
	\subsubsection{Multimodal Video Captioning.}
	Multimodal video captioning aims to generate a sequence of descriptive sentences. As shown in Figure \ref{fig:main_structure} (caption block), the model encodes the input video frames as well as transcripts inside the clips through the video encoder and text encoder respectively, then feeds the encodings to the cross encoder to get unified representation, and finally generates token sequence by the decoder. We use $\mathcal{L}_{Decoder}$ as the loss during the fine-tuning stage. The hyper-parameters are an initial learning rate of 3e-5, a batch size of 32 samples, and fine-tune for 5 epochs.
	
	Table \ref{tab:caption_result} lists the caption results of all baselines and our models on Youcook2. This generation task adopts the corpus-level generation evaluation metric using the pen-source tool\footnote{https://github.com/Maluuba/nlg-eval}, including BLEU (BLEU-3, B-3; BLEU-4, B-4) \cite{papineni2002bleu}, METEOR (M) \cite{banerjee2005meteor}, ROUGE-L (R-L) \cite{lin2004automatic}, and CIDEr \cite{vedantam2015cider}. We compare our pre-trained model with several baseline methods. We classify the methods with the setting that the input is video-only or video+transcript. \citet{zhou2018towards} propose an end-to-end model for both procedural segmentation and captioning.  \citet{sun2019videobert,sun2019contrastive,Zhu_2020_CVPR,Korbar2020} adopt the pre-training strategy and evaluate the captioning with the only video as input. \citet{shi2019dense} and \citet{hessel2019case} discuss the multimodal input with both video and transcript. Our pre-trained model achieves state-of-the-art results and outperforms the existing pre-trained models, even only considering video as input.
	\begin{table*}[tp] 
		\setlength{\tabcolsep}{2pt}
		\centering
		\begin{tabular}{lc}
			\toprule
			Methods         & Frame Accuracy (\%) \\ \midrule
			NN-Viterbi \cite{Richard2018NeuralNetwork}   & 21.17  \\  
			VGG \cite{Simonyan2014Very}  & 25.79  \\  
			TCFPN-ISBA \cite{Ding2018Weakly}   & 34.30  \\  
			CBT  \cite{sun2019contrastive}   & 53.90  \\  
			MIL-NCE \cite{miech19endtoend}   & 61.00  \\  
			ActBERT \cite{Zhu_2020_CVPR}   & 56.95  \\  
			\midrule
			UniVL   & \textbf{70.02} \\
			\bottomrule
		\end{tabular}
		\caption{Action segmentation results on COIN.}
		\label{tab:result_of_segmentation_COIN}
	\end{table*}
	\begin{table*}[tp] 
		\setlength{\tabcolsep}{4pt}
		\centering
		\begin{tabular}{lc}
			\toprule
			Methods         & Average Recall (\%)    \\ \midrule
			\citet{Alayrac2016Unsupervised}   & 13.3 \\  
			\citet{Zhukov2019Cross}   & 22.4 \\  
			Supervised \cite{Zhukov2019Cross}   & 31.6 \\  
			HowTo100M \cite{miech2019howto100m}   & 33.6 \\  
			MIL-NCE \cite{miech19endtoend}   & 40.5 \\  
			ActBERT \cite{Zhu_2020_CVPR}   & 41.4 \\  
			\midrule
			UniVL   & \textbf{42.0 } \\
			\bottomrule
		\end{tabular}
		\caption{Action step localization results on CrossTask.}
		\label{tab:result_of_localization_CrossTask}
	\end{table*}
	
	\subsubsection{Action Segmentation.} 
	We fine-tune our pre-train model on action segmentation task using COIN dataset, which is to predict one pre-defined label for each frame of the given video. As shown in Figure \ref{fig:main_structure} (action tasks block), the model encodes the input video frames through the video encoder, followed by a linear classifier upon the output encodings for frame labeling. We do not use the text encoder due to no text description in the dataset. The evaluation metric is frame-wise accuracy (FA). The hyper-parameters are an initial learning rate of 3e-5, a batch size of 32 samples, and fine-tune for 5 epochs. The results are shown in Table \ref{tab:result_of_segmentation_COIN}. The UniVL significantly outperforms the baselines with more than 14\% improvements. It shows that the pre-trained UniVL actually learns a good visual representation, even absent of linguistic descriptions.
	
	\subsubsection{Action Step Localization.} 
	We evaluate the action step localization on CrossTask dataset. As shown in Figure \ref{fig:main_structure} (action tasks block), the model encodes the step description (action) and video clip through the text encoder and the video encoder respectively. And then calculate the relevance scores through dot product similar to the retrieval task. To fairly compare to \cite{miech2019howto100m,miech19endtoend,Zhu_2020_CVPR}, we do not fine-tune on the CrossTask dataset. We perform the evaluation protocol by reporting the average recall (CTR) metric for the localization task\footnote{The result is generated following the evaluation process of official project: https://github.com/DmZhukov/CrossTask}. The results are shown in Table \ref{tab:result_of_localization_CrossTask}. Our results are even better than the supervised baseline, which demonstrates our UniVL model can learn better joint text-video representation.
	\begin{table*}[tp] 
		\setlength{\tabcolsep}{4pt}
		\centering
		\begin{tabular}{lcccc}
			\toprule
			Methods         & BA & F1 & MAE & Corr \\ \midrule
			MV-LSTM \cite{Rajagopalan2016Extending} & 73.9/- & 74.0/- & 1.019 &  0.601   \\
			TFN \cite{Zadeh2017Tensor} & 73.9/– & 73.4/- & 1.040 & 0.633    \\
			MARN \cite{zadeh2018multi} & 77.1/– & 77.0/- & 0.968 & 0.625    \\
			MFN \cite{zadeh2018memory} & 77.4/– & 77.3/-  & 0.965 & 0.632    \\
			RMFN \cite{Liang2018Multimodal} & 78.4/– & 78.0/- & 0.922 & 0.681    \\
			RAVEN \cite{Wang2019Words} & 78.0/– & -/- & 0.915 & 0.691    \\
			MulT \cite{Tsai2019Multimodal} & –/83.0 & -/82.8 & 0.870 & 0.698    \\
			FMT \cite{Zadeh2019Factorized} & 81.5/83.5 & 81.4/83.5 & 0.837 & 0.744    \\
			\midrule
			UniVL   & \textbf{83.2/84.6} & \textbf{83.3/84.6} & \textbf{0.781} & \textbf{0.767} \\
			\bottomrule
		\end{tabular}
		\caption{Multimodal sentiment analysis results on CMU-MOSI dataset. BA means binary accuracy, MAE is Mean-absolute Error, and Corr is Pearson Correlation Coefficient. For BA and F1, we report two numbers following \citet{Zadeh2019Factorized}: the number on the left side of “/” is calculated based on the approach from \citet{zadeh2018multi}, and the right side is by \citet{Tsai2019Multimodal}.}
		\label{tab:result_of_MOSI}
	\end{table*}
	
	\subsubsection{Multimodal Sentiment Analysis.} 
	We evaluate the multimodal sentiment analysis on CMU-MOSI dataset, the goal of which is to identify the sentiment of speaker based on the speaker’s display of verbal and nonverbal behaviors. We employ video and corresponding transcripts to accomplish this task. As shown in Figure \ref{fig:main_structure} (multimodal classification block), the model encodes the input video frames as well as transcripts inside the clips through the video encoder and text encoder, respectively. Then feeds the encodings to the cross encoder to get unified representation, and finally predicts the sentiment score by a linear on the first token `[CLS]'. The hyper-parameters are an initial learning rate of 1e-5, a batch size 32, and fine-tune for 3 epochs.
	\begin{table*}[tp] 
		\centering
		\setlength{\tabcolsep}{3pt} 
		\begin{tabular}{l|c|cccc}
			\toprule
			Methods  & Dataset & R@1   & R@5 & R@10 & Median R  \\
			\midrule
			UniVL    & Youcook2 & 22.2 & 52.2 & 66.2 & 5 \\
			\multicolumn{1}{r|}{-w/o Joint}   & Youcook2 & 19.5 & 48.0 & 62.7 & 6 \\
			\multicolumn{1}{r|}{-w/o Alignment}   & Youcook2 & 16.3 & 42.3 & 56.2 & 8 \\
			\multicolumn{1}{r|}{-w/o EnhancedV}  & Youcook2 & 16.1 & 41.3 & 55.8 & 8 \\
			\multicolumn{1}{r|}{-w/o Decoder}  & Youcook2 & 14.6 & 40.3 & 55.5 & 8 \\
			\multicolumn{1}{r|}{-w/o StagedP}  & Youcook2 & 11.9 & 35.0 & 48.9 & 11 \\
			\multicolumn{1}{r|}{-w/o Pre-training}  & Youcook2 & 7.7 & 23.9 & 34.7 & 21 \\
			\midrule
			UniVL    & MSR-VTT & 20.6 & 49.1 & 62.9 & 6 \\
			\multicolumn{1}{r|}{-w/o Joint}   & MSR-VTT & 19.6 & 45.9 & 62.6 & 6 \\
			\multicolumn{1}{r|}{-w/o Alignment}   & MSR-VTT & 19.3 & 44.6 & 60.1 & 7 \\
			\multicolumn{1}{r|}{-w/o EnhancedV}  & MSR-VTT & 18.0 & 45.3 & 59.3 & 7 \\
			\multicolumn{1}{r|}{-w/o Decoder}  & MSR-VTT & 18.9 & 44.9 & 57.8 & 7 \\
			\multicolumn{1}{r|}{-w/o StagedP}  & MSR-VTT & 18.0 & 44.3 & 57.7 & 8 \\
			\multicolumn{1}{r|}{-w/o Pre-training}  & MSR-VTT & 16.7 & 44.0 & 55.9 & 8 \\
			\bottomrule
		\end{tabular}
		\caption{Ablation study on retrieval task. `-w/o' means reducing the condition above the previous line.}
		\label{tab:result_of_ablation_retrieval}
	\end{table*}
	
	The results are shown in Table \ref{tab:result_of_MOSI}. Following \cite{Zadeh2019Factorized}, the evaluation metrics are binary accuracy (BA), F1 score, Mean-Absolute Error (MAE), and Pearson Correlation Coefficient (Corr). Compared with the baseline using video, transcript, and audio inputs, our model trained with video and language still achieves the best results without audio information.
	\begin{table*}[tp] 
		\centering
		\setlength{\tabcolsep}{5pt} 
		\begin{tabular}{lccccc}
			\toprule
			Methods  & B-3 & B-4 & M & R-L & CIDEr \\ 
			\midrule
			UniVL & 23.87 & 17.35 & 22.35 & 46.52 & 1.81  \\ 
			\multicolumn{1}{r}{-w/o Joint}  & 23.96 & 17.54 & 22.48 & 46.77 & 1.84   \\ 
			\multicolumn{1}{r}{-w/o Alignment}   & 23.51 & 17.24 & 22.02 & 45.90 & 1.77  \\ 
			\multicolumn{1}{r}{-w/o EnhancedV}  & 23.15 & 17.04 & 21.83 & 45.89 & 1.76 \\ 
			\multicolumn{1}{r}{-w/o Decoder}   & 19.01 & 13.22 & 19.43 & 43.62 & 1.53 \\ 
			\multicolumn{1}{r}{-w/o StagedP}   & 18.13 & 12.49 & 18.78 & 42.64 & 1.46   \\ 
			\multicolumn{1}{r}{-w/o Pre-training} & 14.23	& 9.46	& 16.27	& 37.44	& 1.15 \\
			\bottomrule
		\end{tabular}
		\caption{Ablation study on caption task of Youcook2 dataset. `-w/o' means reducing the condition above the previous line.}
		\label{tab:result_of_ablation_caption}
	\end{table*}
	
	\subsection{Ablation Studies}
	We analyze the effectiveness of our model design on pre-training objectives and strategies through ablation studies over text-based video retrieval and multimodal video captioning tasks. We also discuss the effectiveness of various visual features.
	
	\subsubsection{Modules and Strategies.}
	Table \ref{tab:result_of_ablation_retrieval} shows the effectiveness of each objective or strategy on the retrieval task. The results are reported on both Youcook2 and MSR-VTT datasets. Simultaneously, Table \ref{tab:result_of_ablation_caption} demonstrates the effectiveness of each objective or strategy on the caption task. For the retrieval task, we exploit UniVL (FT-Joint) fine-tuning strategy to study the objectives: Joint loss, Alignment loss, and Decoder loss, and the strategies: StagedP and EnhancedV show consistent improvement. From the result, we can see that the cross encoder and decoder modules can promote the joint representation of video and text. For the caption task, we find that the decoder module shows great advantage and achieves more than 3 points gain on the BLUE-4 metric. Another finding is that the Joint loss decreases the generation task a little, although it performs well in the retrieval task. Excessive emphasis on coarse-grained matching can affect the fine-grained description at the generation task.
	\begin{table}[tp]  \small
		\centering
		\setlength{\tabcolsep}{1.5pt} 
		\begin{tabular}{c|c|cccc}
			\toprule
			Method  & Visual Feature & R@1   & R@5 & R@10 & Median R  \\
			\midrule
			UniVL     & RS152 + RX101 & 11.5 & 29.1 & 40.1 & 17 \\
			on Youcook2    & S3D & 22.2 & 52.2 & 66.2 & 5 \\
			\midrule
			UniVL    & RS152 + RX101 & 18.7 & 44.4 & 58.9 & 7 \\
			on MSR-VTT    & S3D & 20.6 & 49.1 & 62.9 & 6 \\
			\bottomrule
		\end{tabular}
		\caption{Ablation study of visual features for retrieval task. RS152 denotes ResNet-152, RX101 means ResNeXt-101.}
		\label{tab:retrieval_result_visual_feature}
	\end{table}
	\begin{table}[tbp] \small
		\setlength{\tabcolsep}{1.5pt}
		\centering
		\begin{tabular}{l|c|ccccc}
			\toprule
			Method & Visual Feature & B-3 & B-4 & M & R-L & CIDEr \\ \midrule
			\multirow{2}{*}{UniVL} & RS152 + RX101  & 20.42  & 14.31 & 19.92 & 42.35 & 1.47 \\ 
			& S3D  & 23.87 & 17.35 & 22.35  & 46.52 & 1.81   \\  
			\bottomrule
		\end{tabular}
		\caption{Ablation study of visual features for multimodal video captioning results on Youcook2 dataset. RS152 denotes ResNet-152, RX101 means ResNeXt-101.}
		\label{tab:caption_result_visual_feature}
	\end{table}
	
	\subsubsection{Visual Features.}
	\label{sec:visual_features}
	We compare the S3D video feature pre-trained on Howto100M and ResNet-152 plus ResNeXt-101 pre-trained on labeled ImageNet and Kinetics respectively. The ResNet-152 (RS152) and ResNeXt-101 (RX101) are used to extract 2D and 3D features from video clips respectively similar to \citet{miech2019howto100m}'s work. 
	
	As shown in Table \ref{tab:retrieval_result_visual_feature} and Table \ref{tab:caption_result_visual_feature}, the visual feature is important in our pre-training model and the downstream tasks. It is worth studying an end to end training from raw videos instead of extracted fixed video features in the future. However, the time-cost and the memory-cost are enormous. The key bottleneck is visual representation, and we propose two possible approaches: designing a lightweight training scheme, e.g., training on keyframes of video, using a small feature dimension size. 

	\section{Conclusion and Discussion}
	This paper proposes UniVL with self-supervised learning for video and language representation on large scale videos. The UniVL is designed with four modules and five objectives for both video-language understanding and generation tasks. It is a flexible model for most of the multimodal downstream tasks considering both efficiency and effectiveness. We conduct extensive experiments on evaluating our model for five downstream tasks, e.g., text-based video retrieval and multimodal video captioning. The experimental results demonstrate that our pre-trained model can improve the performance to a large extent over the baseline models and achieve state-of-the-art results on five typical multimodal tasks. Besides, we will investigate our model's performance on more massive datasets and more downstream tasks for future work.
	
	\bibliographystyle{acl_natbib}
	\bibliography{egbib}
\end{document}